\documentclass[sigconf]{acmart}
\usepackage{listings}
\usepackage{xcolor}
\usepackage{graphicx}
\usepackage{amsmath}
\usepackage{listings}
\usepackage{float}
\usepackage{floatflt}
\usepackage{comment}
\usepackage{graphicx}
\usepackage{textcomp}
\usepackage{hyperref}
\usepackage{makecell}
\usepackage{tabularx}

\renewcommand\footnotetextcopyrightpermission[1]{} 
\begin{document}
\title{Transfer Learning from One Cancer to Another via Deep Learning Domain Adaptation}

\author{Justin Cheung}
\email{jcheun11@jh.edu}
\orcid{}
\affiliation{%
  \institution{Johns Hopkins University}
  \city{Baltimore}
  \state{Maryland}
  \country{USA}
  }

\author{Samuel Savine}
\email{ssavine1@jh.edu}
\orcid{}
\affiliation{%
  \institution{Johns Hopkins University}
  \city{Baltimore}
  \state{Maryland}
  \country{USA}
  }
  
\author{Calvin Nguyen}
\email{cnguye89@jh.edu}
\orcid{}
\affiliation{%
  \institution{Johns Hopkins University}
  \city{Baltimore}
  \state{Maryland}
  \country{USA}
  }
  
\author{Lin Lu}
\email{llu45@jhu.edu}
\orcid{}
\affiliation{%
  \institution{Johns Hopkins University}
  \city{Baltimore}
  \state{Maryland}
  \country{USA}
  }

\author{Alhassan S. Yasin}
\email{ayasin1@jhu.edu}
\orcid{0009-0001-8033-9850}
\affiliation{%
  \institution{Johns Hopkins University}
  \city{Baltimore}
  \state{Maryland}
  \country{USA}
}

\begin{abstract} 
Supervised deep learning models often achieve excellent performance within their training distribution but struggle to generalize beyond it. In cancer histopathology, for example, a convolutional neural network (CNN) may classify cancer severity accurately for cancer types represented in its training data, yet fail on related but unseen types. Although adenocarcinomas from different organs share morphological features that might support limited cross-domain generalization, addressing domain shift directly is necessary for robust performance. Domain adaptation offers a way to transfer knowledge from labeled data in one cancer type to unlabeled data in another, helping mitigate the scarcity of annotated medical images.

This work evaluates cross-domain classification performance among lung, colon, breast, and kidney adenocarcinomas. A ResNet‑50 trained on any single adenocarcinoma achieves over 98\% accuracy on its own domain but shows minimal generalization to others. Ensembling multiple supervised models does not resolve this limitation. In contrast, converting the ResNet‑50 into a domain adversarial neural network (DANN) substantially improves performance on unlabeled target domains: a DANN trained on labeled breast and colon data and adapted to unlabeled lung data reaches 95.56\% accuracy.

We also examine the impact of stain normalization on domain adaptation. Its effects vary by target domain: for lung, accuracy drops from 95.56\% to 66.60\%, while for breast and colon targets, stain normalization boosts accuracy from 49.22\% to 81.29\% and from 78.48\% to 83.36\%, respectively. Finally, using Integrated Gradients reveals that DANNs consistently attribute importance to biologically meaningful regions such as densely packed nuclei, indicating that the model learns clinically relevant features and can apply them to unlabeled cancer types.

\end{abstract}


\maketitle
\fancyfoot{}
\fancyhead{}
\thispagestyle{empty}

\section{Introduction}
\label{sec:introduction}
Models trained on medical images often fail to generalize when evaluated on data collected under different conditions or anatomical sites. Even within a single cancer type, such as adenocarcinoma, images from various organs show variation in staining, tissue architecture, resolution, and imaging modality \cite{robbins2020cotran}. These differences create a domain shift that can cause a classifier trained on one organ to perform poorly on another despite the same underlying malignant morphology.

Adenocarcinomas across various organs display core histopathological features, such as glandular structures, nuclear atypia, and an increased nucleocytoplasmic ratio, that define this subtype \cite{robbins2020cotran}. These shared characteristics, along with prior observations that metastatic adenocarcinomas may appear similar between tissue sites \cite{dennis2005markers}, raise the possibility that classifiers could transfer information between organ domains.

A second challenge is the imbalance of labeled data across organ sites. Some adenocarcinoma datasets contain large numbers of annotated samples, while others are sparsely labeled due to the cost and time required for expert review \cite{litjens2017survey}. This motivates approaches that can leverage information from well-sampled domains to improve performance in domains with limited representation. Domain adaptation provides one such framework by enabling models to learn domain-invariant representations while still supporting classification.

Evaluating whether cross-organ transfer is reliable, and establishing trust in a deep learning model for this sensitive medical task, requires understanding what image features the model relies on. Explainable AI (XAI) methods such as Integrated Gradients \cite{sundararajan2017axiomatic} provide pixel-level attribution maps that reveal whether predictions are driven by biologically plausible structures such as densely packed nuclei \cite{robbins2020cotran}.

This work evaluates the extent to which adenocarcinoma classifiers trained on one organ site generalize to others and determines whether domain-adaptation strategies improve robustness to domain shift. It furthermore takes an XAI perspective to characterize visual cues used by the model to provide further transparency into how these domain-adapted models make their predictions. In particular, this work demonstrates the value of Integrated Gradients in taking trust in the model's domain adaptation ability beyond we would afford to a purely black-box model; in our experimental results, we demonstrate that an attribution matrix produced by Integrated Gradients aligns well with features in breast cancer histopathology images relevant to physicians for cancer diagnosis, giving visually intuitive evidence that the model learns and leverages biologically meaningful information to perform a meaningful medical task.

\subsection{Related Work}
A number of studies have examined the morphological similarity of adenocarcinomas across tissue sites. Dennis et al. (2005) reported that metastatic adenocarcinomas often appear similar regardless of the organ of origin because the surrounding normal tissue typically used by pathologists to infer tissue site is absent or obscured. Their findings highlight that histologic appearance alone provides limited diagnostic information, motivating the use of molecular or computational methods to distinguish metastatic adenocarcinomas \cite{dennis2005markers}. 

Deep learning studies have also explored cross-cancer generalization. Noorbakhsh et al. (2020) trained convolutional neural networks on 27,815 Cancer Genome Atlas (TCGA) whole-slide images across 19 tumor types and found that models trained on one cancer type could classify others with high cross-cancer AUCs (0.88 $\pm$ 0.11). Their results suggest that conserved morphological patterns across cancers can support generalization and that some cancers (e.g., breast, bladder, uterine) show stronger cross-tissue transferability than others \cite{noorbakhsh2020deep}. Menon et al. (2022) trained ResNet-18 models on 9,792 Cancer Genome Atlas (TCGA) FFPE slides from 7 organs and 11 cancer types, finding that cancers with similar histological structures (e.g., breast, colorectal, liver) showed strong cross organ transfer, whereas kidney and lung cancers displayed more organ-specific morphology and poorer transferability \cite{menon2022exploring}. These findings highlight that similarities in tissue structure play a crucial role in determining when transfer is feasible. 

Domain adaptation methods have been proposed as a way to address imbalances in the availability of labels across datasets. He et al. (2021) developed an adversarial domain adaptation framework in which a feature extractor is trained to produce domain-invariant representations that remain predictive for tumor diagnosis. Using breast tumors as a labeled source domain and gastric tumors as an unlabeled target domain, their method outperformed alternative transfer learning approaches such as an alternative DANN and ADDA on BreakHis and gastric datasets. This demonstrates that adversarial learning can reduce distribution discrepancy and facilitate cross-organ classification without additional labeling \cite{he2021adversarial}. Our work builds on this direction by applying adversarial adaptation to a distinct multi-organ adenocarcinoma dataset and incorporating model explainability.

Another avenue of work focuses on reducing staining and color variability in histology slides. Hoque et al. (2024) performed a comprehensive comparison of stain-normalization techniques, evaluating ten algorithms across multiple datasets. Their study suggests that unified-transformation approaches produced the most consistent and realistic stain profiles and reduced generalization error relative to classical methods such as Reinhard or Macenko. These results highlight the role of stain normalization in mitigating domain differences and improving feature consistency across datasets \cite{hoque2024stain}. In the context of this work, stain normalization can be viewed as a complementary approach to domain adaptation for reducing variation. We hypothesize that stain normalization should help close the domain gap, mitigating the influence of variation in stain techniques across different cancers' datasets in the hope of creating an easier domain adaptation problem wherein a model can just focus on adapting to differences between different organs' cancers from a biological standpoint. In this work, we perform experiments to assess the effect of stain normalization on a DANN's classification accuracy.

Applications of cross-organ cancer modeling continue to materialize in clinical systems. Paige’s AI platform has received Breakthrough Device designation from the U.S. FDA for its ability to detect cancers across a broad range of tissue and organ types directly from histopathology slides. Retamero et al. (2024) evaluated Paige BLN, a deep-learning system trained on over 32,000 whole-slide images, to assess whether AI could enhance the detection of breast cancer lymph-node metastases in routine pathology practice. In a reader study of 167 slide images, the AI assistance increased pathologist sensitivity from 81.2\% to 93.2\% and reduced average review time by more than half, demonstrating that AI can significantly improve both diagnostic accuracy and efficiency. The model itself also achieved high standalone performance with 92.8\% sensitivity and 94.9\% specificity. Together, these findings suggest the feasibility of using well-trained deep learning models to detect cancer directly from histopathology slides for more reliable cancer diagnosis \cite{retamero2024artificial}.

\section{Methods}

\subsection{Dataset}

We conducted our experiments on the Multi Cancer Dataset available on Kaggle \cite{obuli_sai_naren_2024}. This data features 8 cancer types - acute lymphoblastic leukemia (ALL), brain cancer, breast cancer, cervical cancer, kidney cancer, lung and colon cancer, lymphoma, and oral cancer. In the domain adaptation paradigm, we consider each cancer type as a domain. Each cancer type features multiple subclass labels, which differ between cancer types - for example, while the cervical cancer subset of the data includes 5 labels (dyskeratotic, koilocytotic, metaplastic, parabasal, and superficial-intermediate), the kidney cancer subset of the data only includes 2 labels (normal and tumor).
Because the labels are different in both number and meaning between domains, adaptation from one domain to another to transfer learning from the sublabels of one domain to prediction of the sublabels of another domain is not well-defined. For our experimentation, we reduce the problem to ask, \textit{can a cancer classifier distinguish malignant tumors from normal or benign cases}? To answer this question, we binarize the provided labels by grouping them into one of two classes, either normal/benign (0) or malignant (1). Table \ref{tableSubclassLabelBinarization} describes which of these two classes each sublabel from the original dataset corresponds to. The number of samples in the resulting binary classes is indicated in parentheses, in this format: sublabel (number of samples).

\begin{table}[h!]
\caption{Binarization of Provided Cancer Subclass Labels for Domain Adaptation}
\begin{center}
\begin{tabular}{|c|c|c|c|c|c|c|}
\hline
Cancer Type &\multicolumn{2}{|c|}{\textbf{Class Label}} \\
\cline{2-3} 
& \makecell{$y=0$ \\ (Normal/Benign)} & \makecell{$y=1$ \\ (Malignant)}
\\ \hline


breast cancer
& benign (5000)
& malignant (5000)
\\ \hline


kidney cancer
& kidney normal (5000)
& kidney tumor (5000)
\\ \hline

lung cancer
& \makecell{lung benign tissue \\ (5000)}
& \makecell{lung adenocarcinoma \\ (5000)} 
\\ \hline

colon cancer
& \makecell{colon benign tissue \\ (5000)}
& \makecell{colon adenocarcinoma \\ (5000)} 
\\ \hline


\end{tabular}
\label{tableSubclassLabelBinarization}
\end{center}
\end{table}

We further restrict our scope to adenocarcinomas from the Multi Cancer Dataset, leaving four cancer types: breast, kidney, lung, and colon cancers. The available images for breast, lung, and colon cancers are histopathological images, while the kidney cancer images are body CT images. Although CT and histopathology are very different imaging techniques in both scope and interpretation, we leverage the kidney CT images to determine whether the features in.

\begin{figure}[ht!]
    \centering
    \includegraphics[width=1\linewidth]{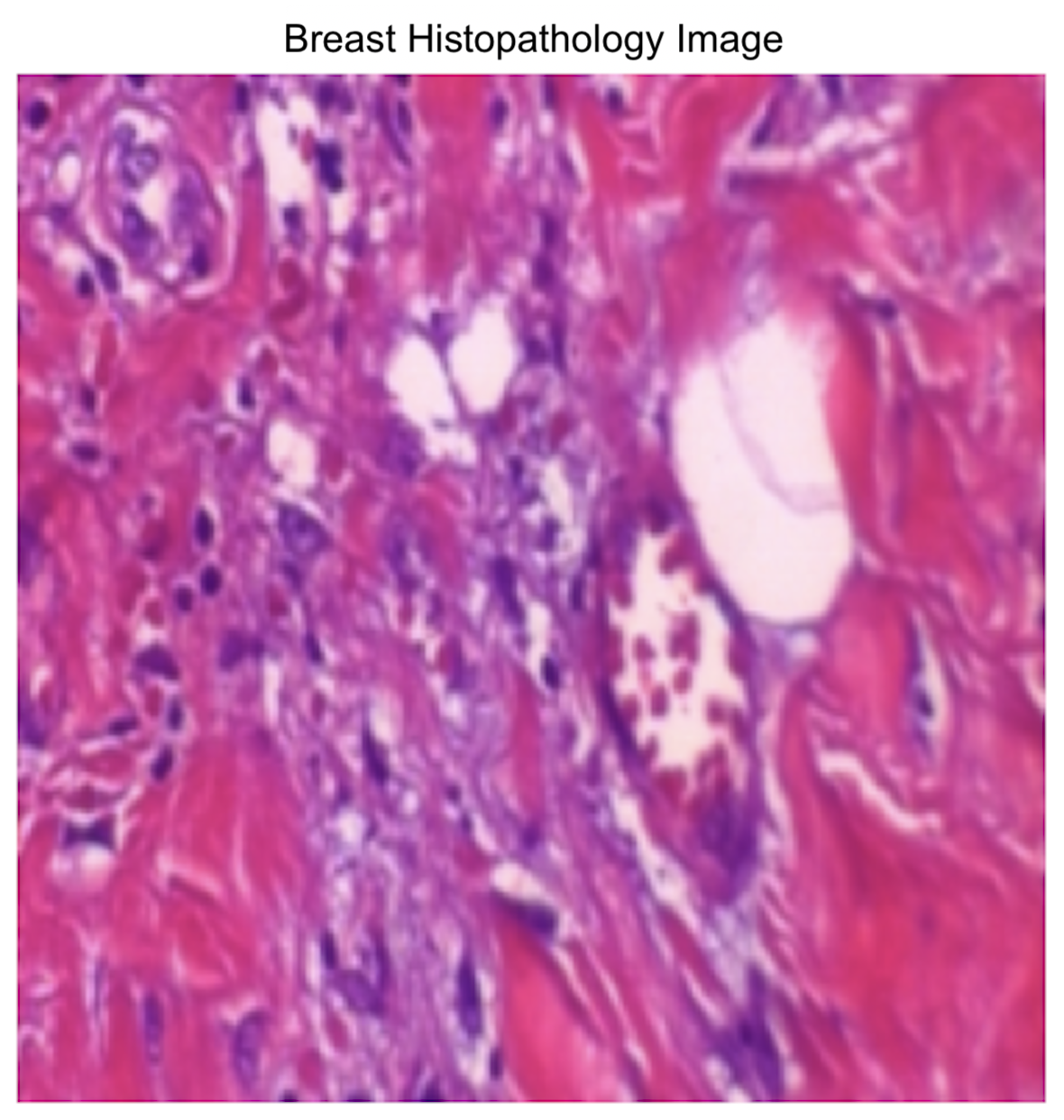}
    \caption{Example of a breast histopathology image. This example is labeled as malignant ($y=1$).}
    \label{figExampleIntegratedGradientsOriginal}
\end{figure}

An example breast cancer histopathology image is shown in Fig.~\ref{figExampleIntegratedGradientsOriginal}.

\subsection{Setup, Training and Evaluation}

\subsubsection{Architecture} 

\begin{figure}[ht!]
  \includegraphics[width=0.9\linewidth]{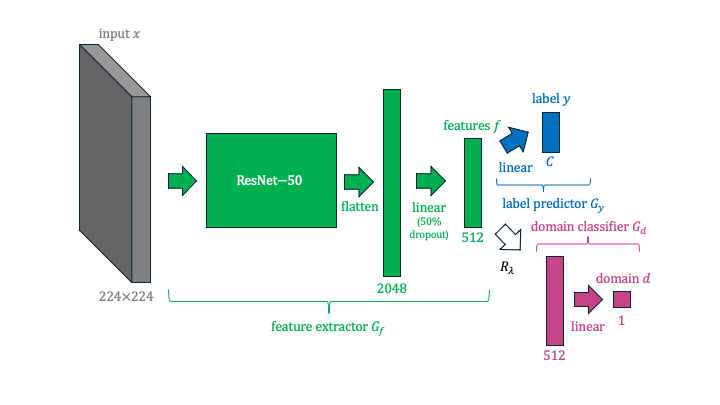}
  \caption{DANN architecture with ResNet-50 feature extractor used in our domain adaptation experiments.}
  \label{figModelArchitecture}
\end{figure}

For our domain adaptation experiments, we use a domain adversarial neural network (DANN) \cite{ganin2015unsupervised}. The architecture we used is illustrated in Fig.~\ref{figModelArchitecture}. We use the same design as in our previous work in CT artifact robustness \cite{cheung2025improvingartifactrobustnessct}, in which ResNet-50 \cite{he2016deep} is used as a feature extractor to learn latent features. These latent features are used both for class label prediction and for domain prediction. In this work, we classify histopathology images into $C=2$ classes: benign/non-cancerous ($y$=0) and malignant ($y$=1). Across our experiments, we investigate different choices of source domain ($d$=0) and target domain ($d$=0), discussed further in the Experimental Design section. Our training hyperparameters are described in Table \ref{tab:hyperparameters}.

\begin{table}[h]
\centering
\caption{Hyperparameters for Domain Adaptation Model Training and Evaluation}
\begin{tabular}{ll}
\hline
\textbf{Parameter} & \textbf{Value} \\
\hline
\multicolumn{2}{l}{\textit{Model Architecture}} \\
Base Model & ResNet-50 \\
Output Feature Dimension & 512 \\
Dropout Rate & 0.5 \\
Number of Classes & 2 \\
\hline
\multicolumn{2}{l}{\textit{Training Configuration}} \\
Optimizer & SGD \\
Learning Rate & 0.1 \\
Weight Decay & 0.0001 \\
Batch Size (Train) & 32 \\
Batch Size (Eval) & 32 \\
Number of Epochs & 50 \\
Max Gradient Norm & 1.0 \\
LR Scheduler & Linear \\
\hline
\multicolumn{2}{l}{\textit{Domain Adaptation}} \\
Lambda (Initial) & 0.0 \\
Lambda Scheduler & Parabolic Increasing \\
Branch 1 Loss & Instance Weighted Cross Entropy \\
Branch 2 Loss & Binary Cross Entropy \\
\hline
\multicolumn{2}{l}{\textit{Data Configuration}} \\
Image Size & $224 \times 224 \times 3$ \\
Data Split Ratio & 70/15/15 (train/val/test) \\
\hline
\end{tabular}
\label{tab:hyperparameters}
\end{table}

\subsubsection{Loss Function}

We implement the loss function used in \cite{ganin2015unsupervised, ganin2016domain}, expressed for a minibatch as in \cite{cheung2025improvingartifactrobustnessct}, as \eqref{eqLoss}:

\begin{multline}
\mathcal{L}(\hat Y, \mathbf{y}, \mathbf{\hat d}, \mathbf{d}) =\\ \sum_{i\in\{1..N\}}(1-d_i)\mathcal{L}_{CE}(\hat y_{i,:}, y_i) + \sum_{i\in\{1..N\}}\mathcal{L}_{BCE}(\hat d_i, d_i) 
\label{eqLoss}
\end{multline}

This loss function can be understood in two parts, corresponding to the dual-branch architecture with distinct loss functions optimized simultaneously through domain adversarial training. The total loss $\mathcal{L}$ combines a classification objective with a domain confusion objective:

\begin{equation}
\mathcal{L} = \mathcal{L}_{\text{class}} + \mathcal{L}_{\text{domain}}
\end{equation}

\paragraph{Classification Loss (Branch 1)}

The primary classification task employs instance-weighted cross-entropy loss to predict binary cancer labels (benign vs. malignant), with masking applied to exclude target domain samples:

\begin{equation}
\mathcal{L}_{\text{class}} = -\frac{1}{N_s} \sum_{i=1}^{N} (1-d_i) \sum_{c=1}^{C} y_{i,c} \log(\hat{y}_{i,c})
\end{equation}

where $N$ is the batch size, $N_s = \sum_{i=1}^{N}(1-d_i)$ is the number of source domain samples, $C=2$ is the number of classes, $y_{i,c}$ is the ground truth label, $\hat{y}_{i,c}$ is the predicted probability for class $c$, and $d_i \in \{0,1\}$ is the domain label (0 for source, 1 for target). The term $(1-d_i)$ effectively masks target domain contributions, allowing only source domain samples with known labels to contribute to the classification loss.

\paragraph{Domain Adversarial Loss (Branch 2)}

The domain classification objective utilizes binary cross-entropy with logits to distinguish between source and target domains:

\begin{equation}
\mathcal{L}_{\text{domain}} = -\frac{1}{N} \sum_{i=1}^{N} \left[ d_i \log(\sigma(z_i)) + (1-d_i) \log(1-\sigma(z_i)) \right]
\end{equation}

where $d_i \in \{0,1\}$ indicates domain membership (0 for source, 1 for target), $z_i$ is the raw logit output, and $\sigma(\cdot)$ is the sigmoid function. This loss operates through a gradient reversal layer (GRL) during backpropagation, which negates gradients by multiplying them by $-\lambda(p)$, where $p = \frac{\text{epoch}}{\text{total\_epochs}} \in [0, 1]$ represents the training progress, and $\lambda$ is the gradient reversal weight. The GRL encourages the feature extractor to learn domain-invariant representations that maximize domain confusion while maintaining classification accuracy on source domain samples.

\subsubsection{Lambda Scheduling}

The gradient reversal weight $\lambda$ follows a parabolic increasing schedule:

\begin{equation}
\lambda(p) = p^2
\end{equation}

This schedule initializes $\lambda(0)=0$ to prioritize learning discriminative features for classification before quadratically increasing domain adaptation pressure, reaching $\lambda(1)=1$ at the end of training to fully prioritize domain adaptation.

We chose this schedule following exploration of a variety of hyperparameter schedules, including a linear increasing schedule ($\lambda(p)=p$), a linear decreasing schedule ($\lambda(p)=-p$), a parabolic decreasing schedule ($\lambda(p)=1-p^2$), a constant schedule ($\lambda(p)=0.5$), and the logistic schedule used in \cite{ganin2015unsupervised} ($\frac{2}{1+\exp{(-10p)}}-1$).


\subsection{Experimental Design} 

\paragraph{Baseline Experiments}

The first experiments we performed were simple baseline experiments to evaluate the architecture's ability to classify samples from each of the domains without attempting to use domain adaptation. These experiments were an attempt to analyze the ability of the resulting model to generalize across the various adenocarcinoma domains. Four models were trained only on their respective domain datasets, but were evaluated on all domain datasets. For example, the breast domain model was only trained on breast images, but was evaluated on the breast, colon, kidney, and lung datasets. This gave us an understanding on whether there was some inherent connection between some of features present within the various adenocarcinoma domains. 

\paragraph{Ensemble Experiments}

After performing the single-domain baseline experiments, we decided to try ensembling the baseline models to see if we can extract additional generalized performance from the baseline models. We created an ensemble model in a leave-one-out style to ensure that the target domain for generalization across domains was not included in the ensembled model. For example, when we tested for generalization to the lung domain, we ensembled the breast, colon, and kidney models. The purpose of this experiment was to see if features learned within the various adenocarcinoma models could be combined to increase domain generalization performance. In all of these cases, the ensemble prediction was calculated as the average of the predicted probabilities of the three models.

\paragraph{Domain Adaptation Experiments}

Similar to our previous work \cite{cheung2025improvingartifactrobustnessct}, we performed domain adaptation experiments that evaluate the domain adaptation capability of the DANN architecture on the adenocarcinoma datasets. In our initial experiment, we included the lung, colon, breast, and kidney images available in the Multi Cancer Dataset. In a follow-up experiment, we excluded the kidney dataset images as they come from a different imaging technique capturing a different scale of anatomy; the kidney images are CT body images in which a full organ can potentially be seen at once, whereas the lung, colon, and breast images are histopathology images which capture cell-level anatomy. For this experiment, we applied target domain labels to the domain we would like to perform domain adaptation towards. These samples' class labels would then be masked by the DANN and would not contribute towards the learning of the model by the mechanism in \eqref{eqLoss} and detaching these samples' class label predictions from the computational graph prior to backpropagation. 

In the next section, we discuss architecture verification experiments which help confirm that the DANN label prediction branch was not able to learn from the target domain samples. However, these target domain samples are used to perform the adversarial gradient reversal step to force the model to learn only the domain invariant features within the adenocarcinoma dataset. 

\paragraph{Architecture Verification} 

During our development, we noticed that gradients were leaking from the domain classifier branch to the label prediction branch. This was unintended in our architecture design, so we performed experiments to verify that the label prediction branch cannot learn on samples from the target domain. For example, if we are trying to adapt to colon images, we cannot have the model's label prediction branch optimize directly from colon images. However, the domain classifier must be able to see these samples and perform gradient reversal in order to force the model to learn only the domain invariant features.

We designed several experiments in order to verify that no gradient leakage occurred within our model. The first experiment we designed was to zero out all of the target labels in a particular dataset (essentially replacing the true labels with erroneous data) and verify that the label prediction branch was not able to properly classify the target domain images. Additionally, we designed an experiment with only the target domain samples. With our architecture the model should not be able to learn on only target domain samples, so the expected classification accuracy is expected to be near random chance (50\% in this case) and to have zero evaluation loss on the label prediction branch which essentially makes the model unable to optimize. These experiments yielded the expected results, building our confidence in our implementation.

\section{Results \& Discussion}

\paragraph{Baseline Experiments}

\begin{figure}[ht!]
    \centering
    \includegraphics[width=1\linewidth]{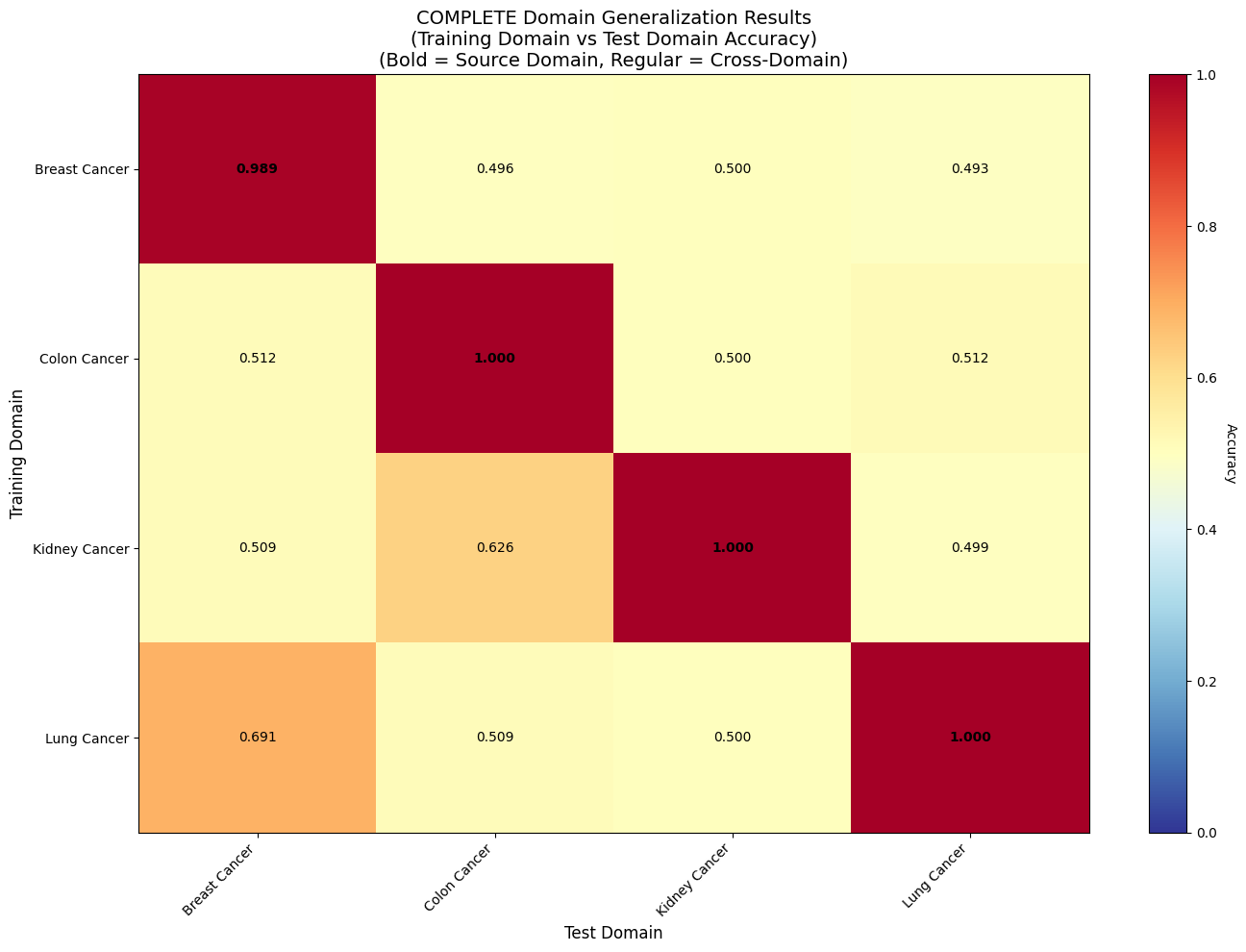}
    \caption{Baseline domain generalization results on the adenocarcinoma dataset}
    \label{baseline_results}
\end{figure}

As we can see in Fig.~\ref{baseline_results}, we can see that in most cases domain generalization was not observed on domains that the model did not directly train on. Note that this model did not perform adversarial domain adaptation to attempt to improve cross domain performance. We observed some minor domain generalization in two cases from this experiment: the model trained on kidney samples displayed some generalization to the colon dataset, and the model trained on lung samples displayed some generalization towards the breast dataset. The kidney model (model trained on kidney samples) achieved 62.6\% classification accuracy on the colon dataset. On these datasets, random chance would be an observed 50\% classification accuracy, which suggests that the kidney model was able to learn some features that allowed it to classify the colon dataset with marginal success. Additionally, the lung model (model trained on lung samples) achieved 69.1\% classification accuracy on the breast dataset. 

\paragraph{Ensemble Experiments}

\begin{figure}[ht!]
    \centering
    \includegraphics[width=1\linewidth]{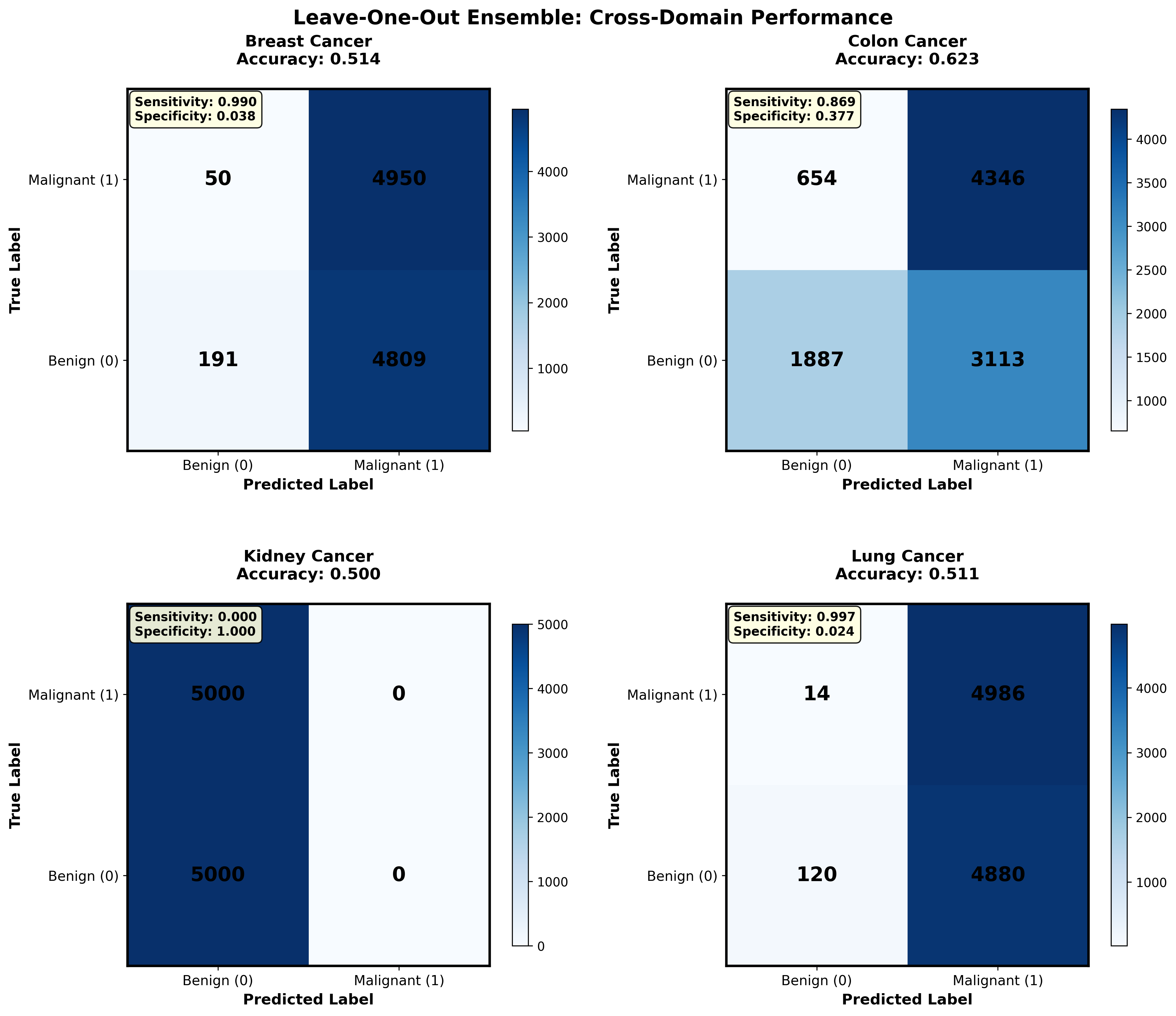}
    \caption{Ensemble model results: titled domain represents the domain that was left out in the leave-one-out ensemble approach and the domain present within the test dataset.}
    \label{ensemble_results}
\end{figure}

From the ensemble results in Fig.~\ref{ensemble_results} we see that the leave-one-out ensemble struggles to adapt towards the unseen domain. Some domain adaptation was observed to the colon domain with a classification accuracy of 62.3\%. The remaining domains achieved classification accuracy close to random chance: 50\%. These results suggest that the breast, kidney, and lung models learned relevant features to increase the resulting ensemble model's ability to classify colon domain images. But, from the baseline results in Fig. ~\ref{baseline_results} we can see that the kidney model achieved a higher classification performance on the colon domain test set. It is likely that the ensemble carried over the colon domain classification ability from the trained kidney model to the ensemble model.

\begin{table}[h]
\centering
\caption{Domain Adaptation Experiment Results}
\begin{tabular}{lllp{0.30\columnwidth}}
\hline
\textbf{Metric} & \textbf{With kidney} & \textbf{No kidney} & \textbf{No kidney+stain normalized} \\

\hline
\multicolumn{4}{l}{\textit{Target Domain: Lung}} \\
Accuracy & 0.8695 & 0.9556 & 0.6660 \\
Precision & 0.8695 & 0.9558 & 0.6771 \\
Recall & 0.8695 & 0.9558 & 0.6676 \\
F1 & 0.8695 & 0.9556 & 0.6619 \\
\hline
\multicolumn{4}{l}{\textit{Target Domain: Breast}} \\
Accuracy & 0.5608 & 0.4922 & 0.8129 \\
Precision & 0.5662 & 0.4936 & 0.8147 \\
Recall & 0.5611 & 0.4943 & 0.8123 \\
F1 & 0.5523 & 0.4795 & 0.8124 \\
\hline
\multicolumn{4}{l}{\textit{Target Domain: Colon}} \\
Accuracy & 0.7515 & 0.7848 & 0.8336 \\
Precision & 0.7671 & 0.7951 & 0.8379 \\
Recall & 0.7518 & 0.7848 & 0.8336 \\
F1 & 0.7480 & 0.7830 & 0.8330 \\
\hline
\multicolumn{4}{l}{\textit{Target Domain: Kidney}} \\
Accuracy & 0.4110 & n/a & n/a \\
Precision & 0.4107 & n/a & n/a \\
Recall & 0.4112 & n/a & n/a \\
F1 & 0.4103 & n/a & n/a \\
\hline
\end{tabular}
\label{tab:DA_results_table}
\end{table}




\paragraph{Domain Adaptation Experiments}

Our original domain adaptation experiments included training the DANN model on the full adenocarcinoma dataset (kidney, lung, breast, colon). The resulting DANN models exhibited significant domain adaptation on the lung and colon domains, slight domain adaptation on the breast domain, and no domain adaptation on the kidney domain. It is likely that the reason the kidney domain wasn't successfully adapted to because of the difference in imaging technique used between the source and target domain datasets. In future experiments we removed the kidney domain and recorded new results, as can be seen in the latter two columns of table~\ref{tab:DA_results_table}. In table~\ref{tab:DA_results_table} first column, we can see that the DANN model that had kidney as the target domain only achieved 41.1\% classification accuracy on the target domain, suggesting that domain adaptation was unsuccessful. We can also see that lung and colon DANN models achieved 86.95\% and 75.15\% classification accuracy on their respective target domain test datasets, which suggests that these domains were successfully adapted to. For the remaining domain, we see slight domain adaptation with a classification accuracy of only 56.08\%, which is not much better than random chance guessing from the DANN model.





For the next domain adaptation experiments, we decided that the kidney domain should be excluded from the adenocarcinoma datasets due to the fact that the imaging technique differed drastically from the rest of the domains. As such we report results for only the breast, colon, and lung domains. As you can see in table~\ref{tab:DA_results_table} second column, we observed significant domain adaptation, where the DANN model adapting to the lung domain achieved 95.56\% classification accuracy and the DANN model adapting to the colon domain achieved 78.48\% classification accuracy. The DANN model adapting to the breast domain did not seem to adapt at all, as it only achieved a classification accuracy of 49.22\% as can be seen in table ~\ref{tab:DA_results_table}. These results are measured from only the target domain samples and do not contain images from the associated domains that the DANN was able to optimize/train to. This suggests that the DANN architecture performs surprisingly well at adapting to certain forms of adenocarcinomas, especially within the lung domain. This is also surprising in comparison to the baseline experiments, as none of the models generalized to the lung domain.

We included a third domain adaptation experiment using a stain-normalized dataset. By performing stain-normalization on the adenocarcinoma dataset we attempted to remove features that might allow the DANN model to learn how to differentiate between the domains by learning the differences in the staining techniques for each domain/sub-domain. This stain-normalized dataset also excludes the kidney domain. As you can see in table~\ref{tab:DA_results_table} third column, the results are generally positive across the domains. But the performance across the domains are also surprisingly different than the dataset without stain-normalization. While the lung domain still exhibits evidence of domain adaptation, it's classification accuracy is only 66.6\%, moving from the highest classification performance of the group towards the lowest. For the breast and colon domains, they achieved 81.29\% and 83.39\% classification accuracy respectively. The breast domain had a drastic improvement of +32.07\% between the two datasets, indicating strong evidence of domain adaptation. The colon domain relatively maintained the same level of performance, increasing by +4.88\%, which is still a significant improvement.

\subsection{Model Explainability}

\begin{figure}[ht!]
    \centering
    \includegraphics[width=1\linewidth]{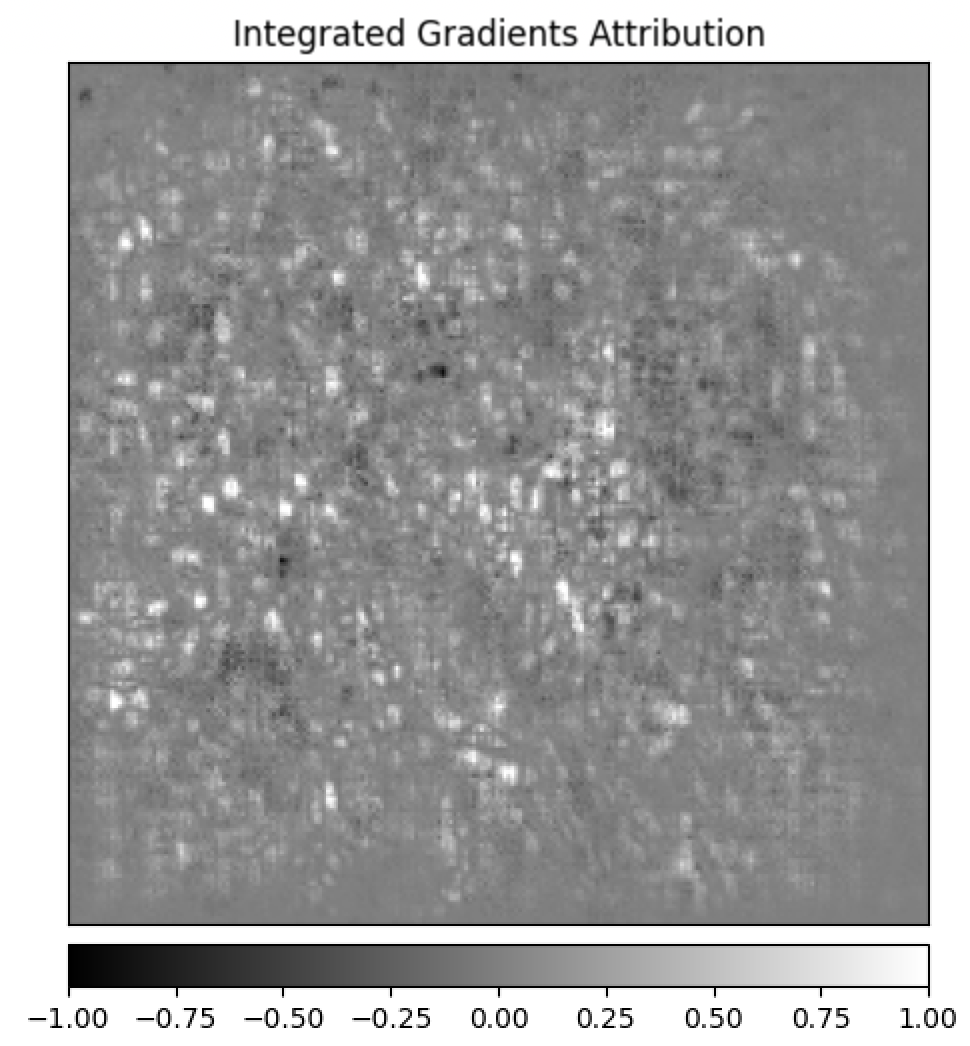}
    \caption{Example of the Integrated Gradients attribution map for the breast target domain DANN for a breast histopathology image from the test set depicted in Fig.~\ref{figExampleIntegratedGradientsOriginal}. The attribution map is colored such that white indicates a positive attribution of a given pixel to the final prediction, while black indicates a negative attribution.}
    \label{figExampleIntegratedGradientsAttributions}
\end{figure}

\begin{figure}[ht!]
    \centering
    \includegraphics[width=1\linewidth]{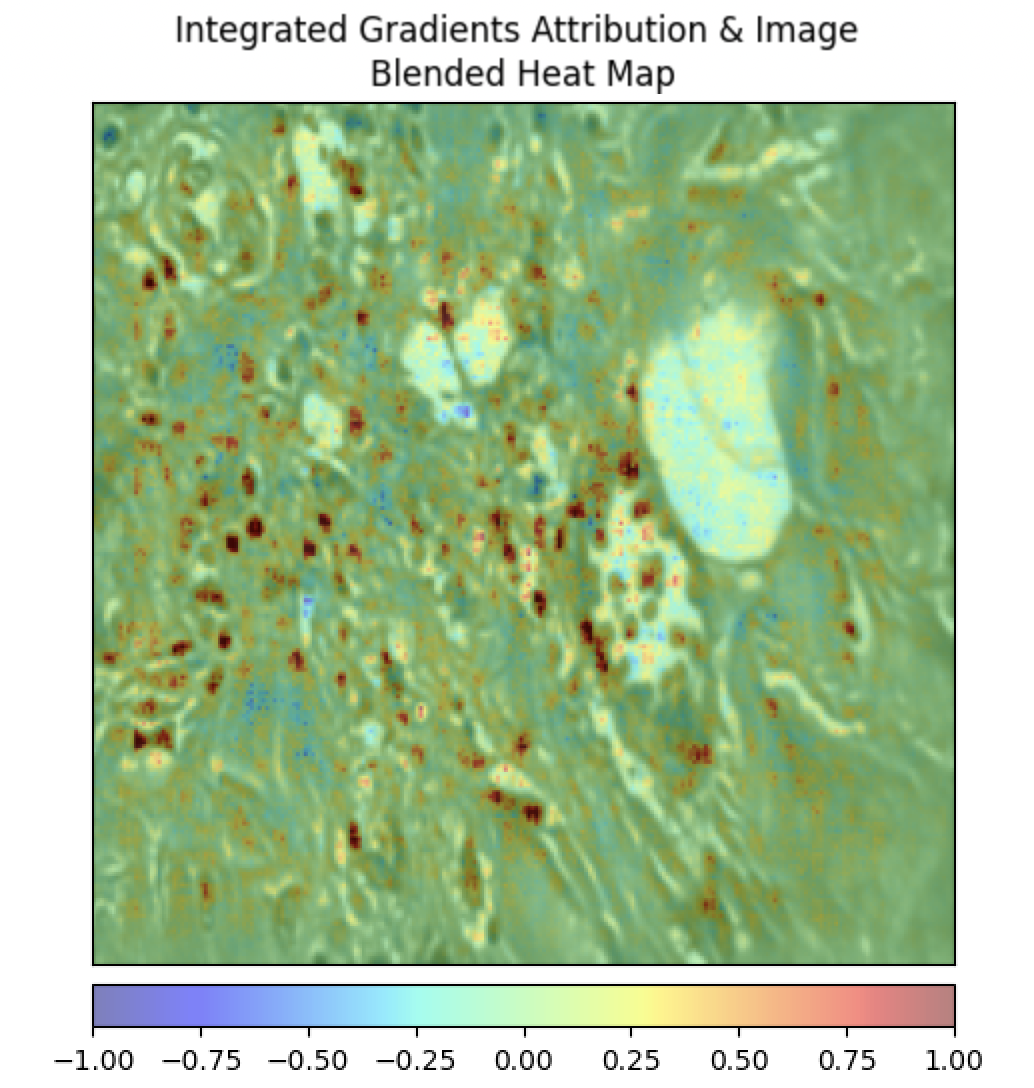}
    \caption{Example of Integrated Gradients attribution map for the breast target domain DANN shown in Fig.~\ref{figExampleIntegratedGradientsAttributions} overlaid onto a breast histopathology image from the test set shown in Fig.~\ref{figExampleIntegratedGradientsOriginal}. The attribution map is colored such that red indicates a positive attribution of a given pixel to the final prediction, while blue indicates a negative attribution. We observe here that the nuclei appear deep red, aligning with our expectation that nuclei which absorb significant amounts of dye and pack densely together are indicative of malignant adenocarcinomas \cite{robbins2020cotran}}
    \label{figExampleIntegratedGradientsBlended}
\end{figure}

For the example histopathology image Fig.~\ref{figExampleIntegratedGradientsOriginal}, we depict an example attribution map generated via Integrated Gradients \cite{sundararajan2017axiomatic} in Fig.~\ref{figExampleIntegratedGradientsAttributions}. We also provide the attribution matrix in color blended with the histopathology image in Fig.~\ref{figExampleIntegratedGradientsBlended} to allow easier visual comparison of the locations in the original histopathology image against locations highlighted by the attribution matrix.

In our usage of Integrated Gradients, we used a baseline image where for a given channel, each pixel in the channel takes the average pixel value for that channel over the training set images. We used contributions from 100 interpolation steps along a linear interpolation path from the baseline image towards a given test image. We note that the result aligns with our expectation that the appearance of dark, densely packed nuclei in a histopathology image is a strong feature to leverage for cancer prediction, as this characteristic of malignant cancer is used by clinicians for diagnosis \cite{robbins2020cotran}.

\section{Conclusion}

Here, we demonstrated that domain adaptation can be used to generalize the cancer classification performance a supervised CNN from one labeled domain (featuring a certain organ and stain protocol) to a new unlabeled domain. In particular, we note that labeled breast and colon adenocarcinoma images alongside unlabeled lung adenocarcinoma images forms a sufficient labeled training dataset for a DANN to learn to classify normal/benign vs. malignant cases for lung adenocarcinoma without the need for lung adenocarcinoma cancer severity labels, at over 95\% accuracy and macro-averaged precision/recall/F1 score. 

We note that ensembling models trained in a purely supervised manner did not help them collectively classify cancer on organs they as a group had not seen images of before, with accuracy around 50.0\%-62.3\% across target domains.

By comparison, domain adaptation via a DANN architecture was far more effective in the case of the models with lung and colon target domains, with target domain accuracy jumping to 86.95\% for the lung target domain model and 75.15\% for the colon target domain model. The breast and kidney target domains remained at 56.08\% and 41.10\% target domain accuracy respectively, suggesting this DANN approach alone was not sufficient to transfer to a new domain.

We also found that mixing imaging techniques by including a kidney adenocarcinoma CT dataset alongside the lung, colon, and breast adenocarcinoma histopathology datasets was generally harmful to the model performance compared to considering only histopathology datasets. The models with lung and colon target domains improved performance when excluding the kidney data from their labeled source domain datasets (improvement from 86.95\% to 95.56\% for lung target domain, and from 75.15\% to 78.48\% for colon target domain), while the model with the breast target domain stayed around random chance of correct classification either way (56.08\% accuracy with kidney included in source domain, 49.22\% accuracy without). This suggests that including the CT dataset, only loosely related in image characteristics to the histopathology datasets, was harmful to adaptation despite providing additional data for an adenocarcinoma classification task.

Finally, we find that stain normalization prior to training greatly affected DANN performance. For the lung target domain, including stain normalization decreased target domain classification accuracy from  95.56\% to 66.60\%. For the other domains, it was helpful - the colon target domain model saw improved accuracy from 78.48\% to 83.36\% with the inclusion of stain normalization. The breast target domain model especially benefited from this normalization, as its accuracy rose from near random chance at 49.22\% up to a meaningful 81.29\% accuracy. These findings suggest that stain normalization can be a useful tool alongside domain adaptation in the adenocarcinoma classification context, further decreasing the aspect of the domain gap related to differences in stain techniques to potentially allow a DANN to more easily focus on other aspects of the domain gap, such as morphological differences between cancer types, for certain cancer types (breast and colon). However, it must be used judiciously, as for lung adenocarcinoma it was harmful to classification performance.

Using Integrated Gradients, we observed that the DANN's performance is explainable, as the model's cancer severity predictions were primarily attributed to cell nuclei visible in a given image, which aligns well with their density and dye absorption being relevant features human experts use for cancer classification \cite{robbins2020cotran}. These findings provide a useful perspective on the ability for deep learning models to exploit similarities among different domains of histopathology images to strengthen their classification performance. Furthermore, they suggest potential for well-founded further trust in deep learning models in oncological tasks.

Our code implementations for this project, including exploration, preprocessing, training, and visualization scripts and Jupyter Notebooks, are available at: \\
\href{https://github.com/JustinCheung168/domain-adaptation-multi-cancer}{https://github.com/JustinCheung168/domain-adaptation-multi-cancer}.

\bibliographystyle{ACM-Reference-Format}
\bibliography{main}

\end{document}